\documentclass[runningheads]{llncs}
\usepackage[T1]{fontenc}

\usepackage{graphicx}
\usepackage{multirow}%
\usepackage{amsmath,amssymb,amsfonts}%
\usepackage{mathrsfs}%
\usepackage[title]{appendix}%
\usepackage{xcolor}%
\usepackage{textcomp}%
\usepackage{manyfoot}%
\usepackage{booktabs}%
\usepackage{algorithm}%
\usepackage{algorithmicx}%
\usepackage{algpseudocode}%
\usepackage{hyperref}
\usepackage{listings}%
\usepackage{xspace}
\usepackage[utf8]{inputenc}
\usepackage{booktabs} % Cho \toprule, \midrule, \bottomrule, \cmidrule
\usepackage{multirow} % Cho \multirow
\usepackage{xcolor}   % Cho các tùy chọn màu sắc, cần cho [HTML]
\usepackage{colortbl} % Cho \cellcolor

\newcommand{\tool}{\textsc{Relabeler}\xspace}
\newcommand{\cola}{\textsc{Cola}\xspace}
\newcommand{\semi}{\textsc{Docta}\xspace}
\newcommand{\semiv}{\textsc{Docta-V}\xspace}
\newcommand{\semir}{\textsc{Docta-R}\xspace}

\begin{document}
\title{A Data-Centric Framework for Detecting and Correcting Corrupted Labels}
\author{Ha-Linh Nguyen \and
Hong-Anh Nguyen \and
Minh-Duc La\and\\
Thu-Trang Nguyen \and
Son Nguyen\and Hieu Dinh Vo\thanks{Corresponding author}}
\authorrunning{Nguyen et al.}

\institute{Faculty of Information Technology,  \\ VNU University of Engineering and Technology, Hanoi, Vietnam\\
\email{\{22024505, 23021466, 23020526, trang.nguyen, sonnguyen, hieuvd\}@vnu.edu.vn}}
\maketitle              % typeset the header of the contribution
%

%%==================================%%
%% Sample for unstructured abstract %%
%%==================================%%
\begin{abstract}
The performance of machine learning and deep learning models largely depends on the quality of the training data. However, the quality of the real-world datasets is often compromised by noisy labels, which can substantially degrade model accuracy and reliability.
To address this challenge, we propose \tool, an end-to-end data-centric framework for detecting and correcting corrupted labels. 
For corrupted label detection, \tool jointly leverages both local and global relationships among data instances to identify potentially noisy samples. 
After detecting suspicious instances, \tool further performs label correction by estimating the most probable clean label for each instance based on both its input features and observed noisy label.
Extensive experiments across multiple datasets, noise types, and noise rates demonstrate that \tool consistently outperforms state-of-the-art baselines, achieving up to 58\% improvement in label correction precision and 6\% improvement in downstream task performance.

\keywords{label noise \and corrupted label detection \and label noise repair \and data-centric}

\end{abstract}

%%================================%%
%% Sample for structured abstract %%
%%================================%%

%%\pacs[JEL Classification]{D8, H51}

%%\pacs[MSC Classification]{35A01, 65L10, 65L12, 65L20, 65L70}

\section{Introduction}
\label{sec:intro}

%importance of clean labelled data and the prevalence of label noise

Machine Learning (ML) and Deep Learning (DL) have achieved impressive performance across various domains~\cite{sharifani2023machine}. Central to the success of ML/DL models is the availability of high-quality labeled training datasets. However, constructing a large-scale and accurately labeled dataset is prohibitively expensive. In practice, labels are often collected via crowd-sourcing or automated annotation tools~\cite{wang2024human}, both of which are inherently error-prone. Consequently, \textit{label noise}, where data instances are assigned incorrect labels, is prevalent, with reported noise rates ranging from 8\% to 38.5\% in real-world datasets~\cite{xiao2015learning}. Such noisy labels can significantly degrade model performance and lead to unreliable predictions~\cite{cola,chenglearning,kim2024learning,sent}. Therefore, effectively handling label noise is essential for developing trustworthy ML/DL systems.

%
%
%Existing Solutions and their Limitations
Existing approaches for handling label noise can generally be divided into two main categories: \textit{model-centric} and \textit{data-centric}.
\textit{Model-centric} approaches focus on improving the robustness of learning algorithms under noisy supervision. These methods typically employ techniques such as robust loss functions~\cite{survey},  regularization strategies~\cite{chenglearning}, or meta-learning frameworks~\cite{li2019learning}, which aim to prevent models from overfitting to incorrect labels.
In contrast, \textit{data-centric} approaches
seeks to improve the dataset quality by identifying and correcting corrupted labels either before or during the training process.
Recent studies in this line of direction~\cite{kim2024learning,cola,simifeat} use statistical patterns, structural assumptions, or dedicated models to detect label errors and enhance data reliability. 

Compared with model-centric methods, data-centric approaches address the root cause of the problem by directly improving the quality of the data itself. 
The improvements are not limited to a specific model, but can benefit any downstream learner trained on the refined dataset. This generality and reusability make data-centric approaches particularly practical for real-world applications.
However, most existing studies~\cite{retrieval-based,yu2023delving,cola,sent} treat noisy label detection and correction as separate tasks and mainly focus on detecting corrupted instances. This separation limits their ability to fully exploit the intrinsic relationship among input features, noisy labels, and clean labels in a unified framework. Furthermore, detecting suspicious instances alone is insufficient, as the remaining challenge is to accurately infer the correct labels for the identified noisy samples.

To address these limitations, this paper proposes \tool, an end-to-end data-centric framework for handling label noise. \tool contains two integrated phases: \textit{corrupted label detection} and \textit{corrupted label correction}. 

In the \textit{corrupted label detection} phase, \tool is motivated by the observation that semantically similar instances are likely to share the same label. Building upon this intuition, \tool jointly leverages both local and global relationships among data instances to identify suspicious ones that are potentially mislabeled. Specifically, suspicious instances are identified as those that exhibit strong similarity while being associated with inconsistent labels.

After detecting suspicious instances, \tool proceeds to the \textit{corrupted label correction} phase, where it estimates the most probable clean label for each instance based on both its input features and its observed noisy label.
In particular, \tool models two complementary factors: how likely an instance belongs to a particular class based on its input features, and how labels are typically corrupted in the dataset.  By integrating these two sources of information through Bayesian inference,  \tool infers the most probable clean label for each suspicious instance.

To evaluate the effectiveness of our proposed framework, we conducted experiments on five widely used datasets covering image, text, and source code classification tasks.
Experimental results under various noise types and noise rates demonstrate that \tool consistently outperforms state-of-the-art methods~\cite{docta}.
In particular, \tool achieves up to 58\% improvement in label correction precision and 6\% improvement in downstream task performance.

The main contributions of this work are summarized as follows:

\begin{itemize}
    \item \tool: an end-to-end data-centric framework for detecting and correcting corrupted labels.
    \item Extensive experiments show that \tool significantly improves label correction precision and reduces error rates, outperforming existing methods.

\end{itemize}

% \section{Label Noise Repair: Problem Formulation}
\section{Problem Formulation}
\label{sec:problem}

Let $\mathcal{X}$ denote the feature space and $\mathcal{Y} = \{1, \dots, C\}$ be the label space. 
A clean dataset is defined as $D = \{(x_i, y_i)\}_{i=1}^{n}$, where each instance $(x_i, y_i)$ is independently drawn from the underlying distribution $\mathcal{D} = \mathcal{X} \times \mathcal{Y}$, and $y_i$ represents the correct label of $x_i$.
However, in real-world scenarios, we typically observe a noisy dataset $\widetilde{D} = \{(x_i, \widetilde{y}_i)\}_{i=1}^{n}$, where each observed label $\widetilde{y}_i$ may differ from the corresponding clean label $y_i$. An instance $(x_i, \widetilde{y}_i)$ is considered corrupted/noisy if $\widetilde{y}_i \neq y_i$, and clean otherwise.
This paper formulates label noise handling as an end-to-end data-centric process consisting of two sequential phases: \textit{corrupted label detection} and \textit{corrupted label correction}.

The goal of \textit{corrupted label detection} is to identify potentially mislabeled instances. Given the noisy dataset $\widetilde{D}$, the detection phase partitions the data into two disjoint subsets $\widetilde{D} = D_{clean} \cup D_{suspicious}$, where $D_{clean}$ contains instances predicted to have reliable labels, while $D_{suspicious}$ contains suspicious instances identified as potentially mislabeled.

Subsequently, the \textit{corrupted label correction} phase aims to repair labels of instances in $D_{suspicious}$. 
Specifically, for each suspicious instance $(x_j, \widetilde{y}_j) \in D_{suspicious}$, the objective of this phase is to infer a corrected label $y_j^*$ that best approximates the unknown clean label $y_j$. The repaired dataset is then constructed as:
$$
D_{repaired} = \{(x_j, y_j^*) | (x_j, \widetilde{y}_j) \in D_{suspicious}\}
$$

Finally, the repaired instances can be combined  with $D_{\text{clean}}$ to construct a high-quality dataset for downstream model training: $D_{final} = D_{clean} \cup D_{repaired}$.

\section{\tool: A Data-Centric Framework for Detecting and Correcting Corrupted Labels}
\label{sec:approach}

\begin{figure*}
    \centering
    \includegraphics[width=\textwidth]{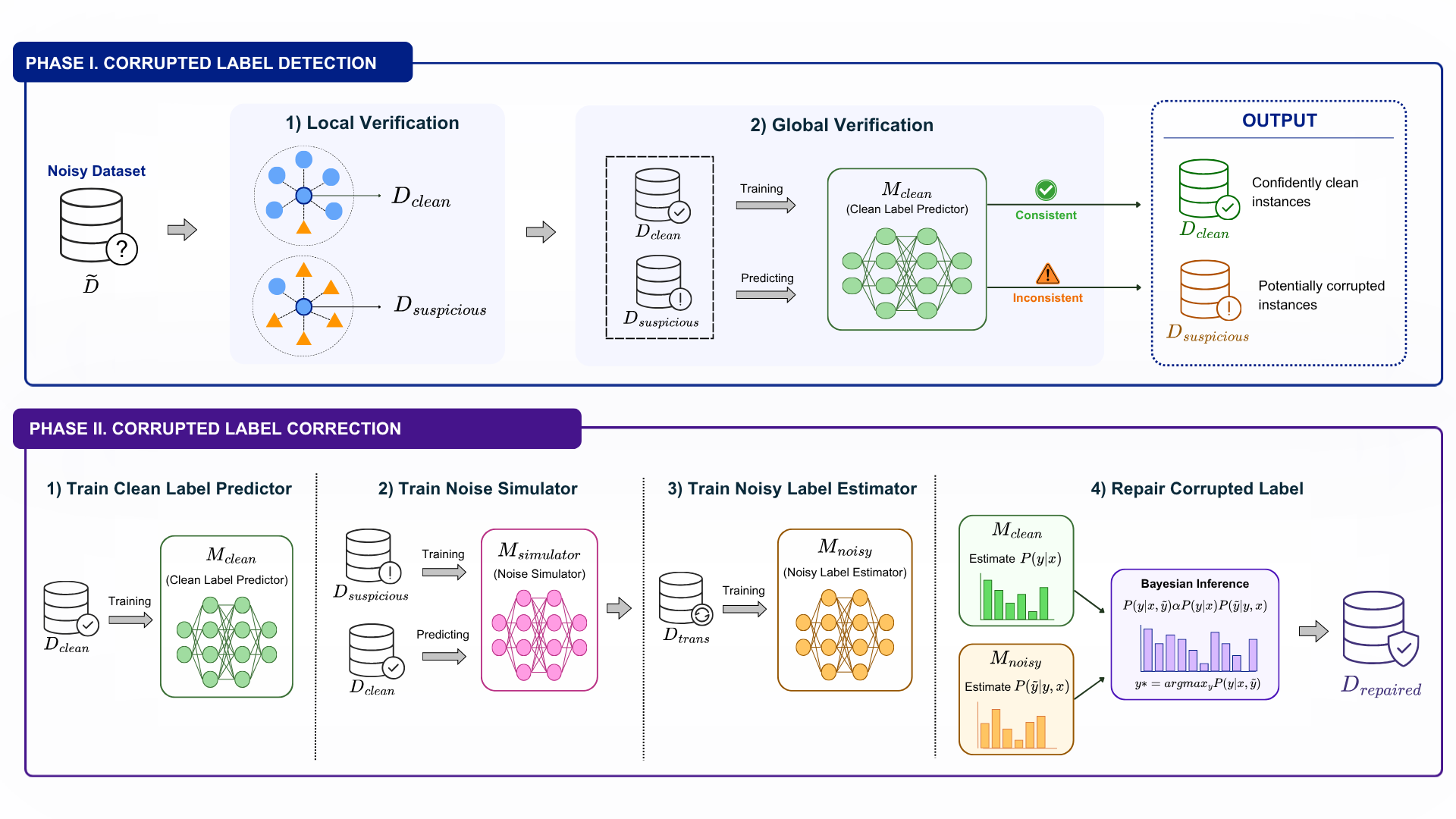}
    \caption{\centering The overall pipeline of \tool}
    \label{fig:pipeline}
\end{figure*}

% \subsection{Approach Overview}

Fig.~\ref{fig:pipeline} shows the overview of \tool, which consists of two main phases: (i)~\textit{corrupted label detection} and (ii)~\textit{corrupted label correction}. Given a noisy dataset $\widetilde{D}$, Phase~1 identifies potentially mislabeled instances and partitions $\widetilde{D}$ into two subsets: a clean subset $D_{clean}$ and a suspicious subset $D_{suspicious}$. Phase~2 then takes $D_{clean}$ and $D_{suspicious}$ as input to repair the labels of suspicious instances in $D_{suspicious}$. The output of Phase~2 is a repaired dataset $D_{repair}$, in which corrupted labels in $D_{suspicious}$ are corrected to their most probable clean labels.

\subsection{Corrupted Label Detection}

This phase aims to detect potentially mislabeled instances by identifying semantically similar instances that are associated with inconsistent labels. The underlying intuition is that \textit{instances with similar features are likely to share the same label.} Therefore, when similar instances are assigned different labels, some of them are likely to be mislabeled. 
To comprehensively verify labels, \tool examines data instances from both \textit{local} and \textit{global} similarity perspectives.

First, \tool performs \textit{\textbf{local verification}} by examining the labels of nearby instances in the feature space. Specifically, for instance $x_i$, \tool verified whether its label $\widetilde{y}_i$ is consistent with those of its nearby neighbors. To achieve this, \tool employs the $k$-Nearest Neighbors ($k$-NN) algorithm to retrieve the $k$ nearest neighbors of $x_i$. $k$-NN is adopted because it is widely used for local similarity analysis and has been shown to be relatively robust to label noise~\cite{gao2016resistance}. 
If the label of $x_i$ is consistent with the majority label among its neighbors, the instance is considered likely to be clean and is added to $D_{clean}$. Otherwise, it is regarded as suspicious and added to $D_{suspicious}$.

Although $k$-NN is effective in capturing local similarity, it only considers a limited neighborhood around each instance. 
Consequently, the local verification step can achieve high precision in identifying clean instances.
However, it may incorrectly classify clean instances as suspicious, since an instance whose label differs from those of its nearby neighbors may still be clean when viewed from a broader global context. Such cases commonly arise near class boundaries or in regions with complex data distributions. 
To address this limitation, \tool performs an additional \textit{global verification} step to further examine suspicious instances in a global similarity perspective.

For \textit{\textbf{global verification}}, \tool uses $D_{clean}$ to train a classification model $\mathcal{M}_{clean}$. 
Since $D_{clean}$ only contains likely clean instances, the resulting model is expected to provide reliable predictions.
More importantly, $D_{clean}$ can contain multiple local regions, enabling $\mathcal{M}_{clean}$ to capture broader and more general patterns across the entire dataset. For each instance $x_i \in D_{suspicious}$, if the label predicted by $\mathcal{M}_{clean}$ matches its original label, the label of $x_i$ is consistent with global patterns captured by $\mathcal{M}_{clean}$.
In this case, $x_i$ is considered to be likely clean and is moved to $D_{clean}$. Otherwise, the instance remains in $D_{suspicious}$ as a potentially corrupted sample. This process is performed iteratively. After each iteration, newly verified clean instances are added to $D_{clean}$, enabling the classification model to be progressively refined using increasingly reliable training data. The iterative process continues until no additional clean instances can be identified.

\textit{In summary}, the overview of the corrupted label detection phase is shown in the top half of Fig.~\ref{fig:pipeline}. This phase 
takes the noisy dataset $\widetilde{D}$
as input and performs two verification steps. First, \tool conducts local verification to identify likely clean samples whose labels are consistent with those of their similar instance in the local context. Next, \tool performs global verification to further identify likely clean samples whose labels are consistent with broader global patterns across the dataset.
The output of this phase is a partition of $\widetilde{D}$ into two subsets: a clean subset $D_{clean}$ and a suspicious subset $D_{suspicious}$.

\subsection{Corrupted Label Correction}

The corrupted label correction phase takes as input two disjoint sets obtained from the previous phase, $D_{clean} = \{(x_i, {y}_i)\}_{i=1}^m$ and $D_{suspicious} = \{(x_j, \tilde{y}_j)\}_{j=1}^{m'}$. The goal of this phase is to
\textit{repair corrupted labels in $D_{suspicious}$ by estimating the most probable correct label for each instance given both its input features and its observed noisy label.} Specifically, for an instance $(x, \tilde{y}) \in D_{suspicious}$, \tool estimates the repaired label $y^*$ by maximizing the posterior probability $P(y \mid x, \tilde{y})$ as defined in Eq.~\ref{eq:repaired_label_estimation}: 
\begin{equation}
    y^* = \underset{y \in \mathcal{Y}}{\arg\max} P(y \mid x, \tilde{y})
    \label{eq:repaired_label_estimation}
\end{equation}
Applying Bayes' theorem, Eq.~\ref{eq:repaired_label_estimation} can be reformulated as Eq.~\ref{eq:bayes_applied}. The detailed derivation of this formulation is available on our website~\cite{website}.
\begin{equation}
    y^* = \underset{y \in \mathcal{Y}}{\arg\max} \left(  {P(y \mid x)} \cdot {P(\tilde{y} \mid x, y)} \right)
    \label{eq:bayes_applied}
\end{equation}
%
%
% combines two key components: ${P(y| x)}$ and $P(\tilde{y}|y, x)$.  
Eq.~\ref{eq:bayes_applied} combines two complementary sources of information. The first term, ${P(y \mid x)}$ estimates how likely $y$ is the correct label based on the input feature $x$. The second term, $P(\tilde{y} \mid x, y)$, models the probability of observing the noisy label $\tilde{y}$, given the input feature $x$ and a candidate label $y \in \mathcal{Y}$.
To measure these probabilities, \tool introduces two models: a \textit{{clean label predictor}}  $\mathcal{M}_{clean}$ and a \textit{{noisy label estimator}} $\mathcal{M}_{noisy}$.

First, to model the distribution ${P(y \mid x)}$, \tool trains a \textit{\textbf{clean label predictor}} $\mathcal{M}_{clean}$ using the subset $D_{clean}$. Since $D_{clean}$ contains instances whose labels are likely correct, it enables  $\mathcal{M}_{clean}$  to learn the underlying relationship between the input features and clean labels. 
Second, to approximate the noise transition probability $P(\tilde{y} \mid x, y)$, \tool trains a \textit{\textbf{noisy label estimator}} $\mathcal{M}_{noisy}$. This model takes both instance $x$ and a candidate label $y$ as input, and predicts the likelihood of observing the noisy label $\tilde{y}$. 

Training $\mathcal{M}_{noisy}$ requires transition data in the form of triples $((x, y), \tilde{y})$ that explicitly characterize how clean labels are corrupted into noisy ones. 
However, such transition data is not directly available. To address this issue, we construct a transition dataset $D_{{trans}} = \{((x_i, y_i), \tilde{y}_i) \mid (x_i, y_i) \in D_{clean}\}$ by injecting synthetic noise to $D_{clean}$ in a realistic manner.

To generate realistic noisy labels, \tool trains a separate neural network called \textit{\textbf{noise simulator}}  $\mathcal{M}_{{simulator}}$ using the suspicious subset $D_{suspicious}$. Since $D_{suspicious}$ contains instances that are likely to be corrupted, training on this subset enables $\mathcal{M}_{simulator}$ to learn the statistical relationship between input features and the noisy labeling patterns present in the dataset.
After training, $\mathcal{M}_{{simulator}}$ is applied to instances in $D_{clean}$ to generate synthetic noisy labels. 
Specifically, for each instance $(x_i, y_i) \in D_{clean}$, the simulator predicts a noisy label $\tilde{y}_i = \mathcal{M}_{{simulator}}(x_i)$, forming a triplet $(x_i, y_i, \tilde{y}_i)$. These triplets are then used to construct the transition dataset $D{_{trans}}$ for training the noisy label estimator.

Using the transition dataset $D_{trans}$, we train  $\mathcal{M}_{noisy}$ to learn the noise transition function. By training $\mathcal{M}_{noisy}$ on $D_{{trans}}$, the model learns to capture how a clean label $y$ might be mislabeled to $\tilde{y}$ given the feature $x$. For each input feature $x$, $\mathcal{M}_{noisy}$ is designed to output a $C \times C$ transition matrix $T_x$, where $C$ is the number of classes in the label space $\mathcal{Y}$.
Each element $T_x[l,k]$ represents the conditional probability $P(\tilde{y}=k \mid x, y=l)$. 
This is the likelihood of observing a noisy label $\tilde{y}=k$ given input $x$ and its candidate clean label $y = l$.

The workflow of the corrupted label correction phase is illustrated in the bottom half of Fig.~\ref{fig:pipeline}. 
First, the clean label predictor $\mathcal{M}_{clean}$ is trained on $D_{clean}$. In \tool, $\mathcal{M}_{clean}$ can reuse the classification model obtained from the corrupted label detection phase.
Next, \tool trains the noise simulator $\mathcal{M}_{simulator}$ on $D_{suspicious}$ for simulating noisy labels and constructing the transition dataset $D_{trans}$. 
Using $D_{trans}$, \tool subsequently trains the noisy label estimator $\mathcal{M}_{noisy}$. 
Finally, given a suspicious instance $(x, \tilde{y})$, \tool employs  $\mathcal{M}_{clean}$ and $\mathcal{M}_{noisy}$ to compute $P(y \mid x)$ and  $P(\tilde{y} \mid x, y)$ for each candidate label $y \in \mathcal{Y}$. These estimates are integrated using the formulation defined in Eq.~\ref{eq:bayes_applied}, the repaired label $y^*$ is the label with the highest posterior probability.

Although the label with the highest probability $P(y \mid x)$ predicted by $\mathcal{M}_{clean}$ could be directly used to repair the label of $x$, relying on it alone is insufficient and potentially unreliable. While $\mathcal{M}_{clean}$ can capture the general relationship between features and clean labels, it does not account for the noise distribution in the dataset. 
By integrating predictions from both $\mathcal{M}_{clean}$ and $\mathcal{M}_{noisy}$, our method incorporates both the semantic alignment of features with clean labels and the statistical patterns of label corruption. This combination enables more informed and context-aware label repair. The effectiveness of this combination is empirically evaluated in Sec.~\ref{sec:ablation_study}.

\section{Experimental Results}
\label{sec:results}

\subsection{Research Questions and Experimental Setup}
To comprehensively evaluate the effectiveness of \tool in detecting and correcting corrupted labels, we seek to answer the following research questions:

\begin{itemize}
    \item \textbf{RQ1. Performance Comparison:} How effective is \tool in detecting and repairing noisy labels? How does it compare with state-of-the-art approaches~\cite{simifeat}?
    \item \textbf{RQ2. Robustness to Noise Types:} How does \tool perform under different types of label noise?
    \item \textbf{RQ3. Component Analysis:} What are the contributions of the key components of \tool to the overall performance?
 
    \item \textbf{RQ4. Downstream Utility:} To what extent do datasets corrected by \tool improve the performance of the downstream models?
\end{itemize}

\begin{table}[!htp]\centering
\caption{Dataset Overview}
\label{tab:dataset}

\begin{tabular}{c|c|r|rr}\toprule
\textbf{Dataset} &\textbf{Data type} &\textbf{\#Instances} &\textbf{\#Labels} \\\midrule
CIFAR-10 &image &50K &10 \\
CIFAR-100 &image &50K &100 \\
Agnews &text &127.6K &4 \\
Clickbait & text & 32K&2\\
CodeXGLUE &code & 27.3K & 2\\
\bottomrule
\end{tabular}
\end{table}

Our experiments are conducted on five widely-used datasets that are commonly adopted in existing data-centric studies~\cite{kim2024learning,cola,simifeat}, including \textit{CIFAR-10}, \textit{CIFAR-100} for image classification, \textit{Agnews} and \textit{Clickbait} for text classification, and \textit{CodeXGLUE} for source code classification. These datasets are diverse in tasks, domains, and the number of labels. Table~\ref{tab:dataset} summarizes the overview of the datasets.

To evaluate the effectiveness of \tool, we compare it against the state-of-the-art approach, \semi~\cite{docta}, which includes two variants:  \textit{voting-based} approach, \semiv, and \textit{ranking-based} approach, \semir.  
In particular, \semi detects and repairs corrupted labels by examining the label consistency among neighboring instances.
For each detected corrupted instance, \semiv assigns the majority-voted label as the repaired label, while \semir selects the label with the highest ranking score.  
To ensure a fair comparison, all methods use BERT~\cite{bert} to encode text data and CLIP~\cite{clip} to encode image data.

We assess the performance of \tool and both variants of \semi using two metrics: \textbf{Precision} and \textbf{Error Rate}.
\textit{Precision} measures the accuracy of corrections, defined as the proportion of correctly changed labels among all changed labels; higher values indicate better performance.
\textit{Error Rate} quantifies the proportion of incorrect labels remaining in the dataset after the correction process; lower values indicate better performance.

\subsection{Experimental Results}

\subsubsection{RQ1: Performance Comparison}
\label{sec:performance_comparison}

% \begin{table*}\centering
% \caption{\centering RQ1: Label noise repair performance of the approaches on different datasets}
% \label{tab:rq1}
% \begin{tabular}{l|l|rr|rr|rr|rr|rr}\toprule
% \multirow{2}{*}{} & &\multicolumn{2}{|c|}  {\textbf{Agnews}} &\multicolumn{2}{|c|}{\textbf{CIFAR-10}} &\multicolumn{2}{|c}{\textbf{CIFAR-100}} & \multicolumn{2}{|c}{\textbf{Clickbait}} & \multicolumn{2}{|c}{\textbf{CodeXGLUE}}\\\midrule
% &Noise rate (\%) &\textbf{20.00} &\textbf{40.00} &\textbf{20.00} &\textbf{40.00} &\textbf{20.00} &\textbf{40.00} &\textbf{20.00} &\textbf{40.00} &\textbf{20.00} &\textbf{40.00}\\ \midrule
% \multirow{2}{*}{\textbf{\semiv}} &Precision (↑) &64.72 &56.70 &70.10 &61.94 &30.49 &34.09 & 85.38 & 66.93\\
% &Error rate (↓) &11.31 &31.6 &8.40 &24.75 &32.57 &37.14 & 3.72 & 22.32\\\midrule
% \multirow{2}{*}{\textbf{\semir}} &Precision (↑) &73.15 &61.43 &79.25 &67.50 &47.63 &40.78 & 85.38 & 68.53\\
% &Error rate (↓) &11.33 &29.73 &8.40 &23.54 &32.52 &47.89 & 3.72 & 21.71\\\midrule
% \multirow{2}{*}{\textbf{\tool}} &Precision (↑) &77.41 &89.47 &84.64 &91.28 &37.88 &45.78 & 96.12 & 97.76\\
% &Error rate (↓) &6.78 &14.00 &4.01&7.48 &20.26 &34.46 & 1.68 & 5.87\\
% \bottomrule
% \end{tabular}
% \end{table*}

%Please add the following packages if necessary:
%\usepackage{booktabs, multirow} % for borders and merged ranges
%\usepackage{soul}% for underlines
%\usepackage{xcolor,colortbl} % for cell colors
%\usepackage{changepage,threeparttable} % for wide tables
\begin{table}[!htp]\centering\small
\caption{\centering RQ1: Label noise repair performance of the approaches on different datasets}
\label{tab:rq1}
\resizebox{\textwidth}{!}{ 
\begin{tabular}{l|r|r|r|r|r|r|rr}\toprule
& &\multicolumn{2}{c|}{\textbf{\semiv}} &\multicolumn{2}{c|}{\textbf{\semir}} &\multicolumn{2}{c}{\textbf{\tool}} \\\cmidrule{3-8}
&Noise rate (\%) &\textbf{Precision (↑)} &\textbf{Error rate (↓)} &\textbf{Precision (↑)} &\textbf{Error rate (↓)} &\textbf{Precision (↑)} &\textbf{Error rate (↓)} \\\midrule
\multirow{2}{*}{Agnews} &20 &64.72 &11.31 &73.15 &11.33 &77.41 &6.78 \\
&40 &56.70 &31.60 &61.43 &29.73 &89.47 &14.00 \\\midrule
\multirow{2}{*}{Clickbait} &20 &85.38 &3.72 &85.38 &3.72 &96.12 &1.68 \\
&40 &66.93 &22.32 &68.53 &21.71 &97.76 &5.87 \\\midrule
\multirow{2}{*}{CIFAR-10} &20 &70.10 &8.40 &79.25 &8.40 &84.64 &4.01 \\
&40 &61.94 &24.75 &67.50 &23.54 &91.28 &7.48 \\\midrule
\multirow{2}{*}{CIFAR-100} &20 &30.49 &32.57 &47.63 &32.52 &37.88 &20.26 \\
&40 &34.09 &37.14 &40.78 &47.89 &45.78 &34.46 \\\midrule
\multirow{2}{*}{CodeXGLUE} &20 &23.27 &42.61 &23.57 &38.91 &33.14 &19.85 \\
&40 &40.69 &45.73 &40.58 &44.66 &52.87 &36.44 \\
\bottomrule
\end{tabular}
}
\end{table}

Table~\ref{tab:rq1} shows the performance of label noise handling approaches across datasets under different rates of instance-dependent noise. In this setting, the corruption probability depends on the intrinsic characteristics of each instance.
This noise type is realistic and challenging, as mislabeled instances often have ambiguous or hard-to-distinguish features.
Overall, \textit{\tool consistently outperforms both variants of \semi in terms of both Precision and Error Rate.}

Specifically, \tool is able to detect and correct noisy labels more accurately than the baselines. It achieves improvements ranging from 6\% to 58\% across different datasets and noise levels. For instance, on the \textit{Agnews} dataset with a noise rate of 40\%, \tool correctly repairs 90\% of the detected corrupted labels, while \semiv achieves only 57\%.
These results highlight the effectiveness of \tool in high-noise scenarios.

Additionally, \tool more effectively reduces the remaining noise after label correction. It lowers the residual noise rate by 3\% to 17\% in absolute terms compared to the baselines. For example, on \textit{CIFAR-10} with an initial noise rate of 40\%, \tool reduces the error rate to just 7\%, whereas \semiv and \semir retain much higher post-correction noise rates of 25\% and 24\%, respectively. This result indicates that \tool can infer more reliable clean labels and substantially improve overall dataset quality.

In summary, the experimental results demonstrate that \tool consistently achieves more accurate corrupted label detection and more effective noise removal than the baseline methods. These findings highlight the robustness of \tool and confirm its effectiveness across diverse datasets and experimental settings.

\subsubsection{RQ2: Robustness to Noise Types}
\begin{table}
\centering
\caption{\centering Label noise repair performance under different noise types}\label{tab:rq2}
%\resizebox{ extwidth}{!}{ % use this if the table is too large
\begin{tabular}{l|l|r|r|r|rr}\toprule
\textbf{} &\textbf{Noise type} &\textbf{Inst.} &\textbf{Symm.} &\textbf{Asym.} & \textbf{Mix.} \\\midrule
\multirow{2}{*}{\textbf{\semiv}} &Precision (↑) &61.94 &78.86 &62.25 &76.38 \\
&Error rate (↓) &24.75 &10.21 &25.71 &23.62 \\\midrule
\multirow{2}{*}{\textbf{\semir}} &Precision (↑) &67.50 &84.62 &64.21 &83.62 \\
&Error rate (↓) &23.54 &10.21 &28.28 &16.38 \\\midrule
\multirow{2}{*}{\textbf{\tool}} &Precision (↑) &91.28 &90.26 &95.37 &88.20 \\
&Error rate (↓) &7.48 &5.01 &9.87 &11.80 \\
\bottomrule
\end{tabular}
\end{table}

To evaluate the robustness of the approaches under different noise types~\cite{cola,simifeat}, we conduct experiments on the CIFAR-10 dataset, where each setting includes only one specific type of noise, including \textit{Symmetric} (\textit{Symm.}), \textit{Asymmetric} (\textit{Asym.}), or \textit{Instance-dependent} (\textit{Inst.}), or a mixture of all types (\textit{Mix.}).
In particular, \textit{\textit{Symm.}} noise randomly flips labels to any other class with equal probability, representing class-agnostic and unbiased corruption. \textit{Asym.} noise introduces class-dependent label flips that follow semantic similarities between classes, modeling systematic and structured annotation errors. \textit{Inst.} noise assigns corrupted labels based on instance-specific characteristics, making the noise pattern dependent on input features and therefore more challenging to detect and repair. Finally, the \textit{Mix.} setting combines all three noise types to simulate realistic and heterogeneous label corruption commonly observed in real-world datasets.

As shown in Table~\ref{tab:rq2}, \textit{\tool exhibits superior robustness and effectiveness across all scenarios}. Specifically, \tool consistently achieves the highest Precision and the lowest Error Rate for every noise type.
For example, under \textit{Asym.} noise, \tool obtains a label repair Precision of 95.37, and reduces the post-correction Error Rate to 9.87.
This corresponds to relative improvements of  49\% and 53\% in Precision and 65\% and 62\% in Error Rate compared to \semir and \semiv, respectively.

Both \semiv and \semir perform reasonably well under \textit{Symm.} noise; however, their performance decreases significantly under settings of more challenging noises, such as \textit{Asym}. and \textit{Inst.}.
Particularly, \semiv and  \semir achieve post-correction Error Rates of around 10\% under the setting of \textit{Symm.} noise. Meanwhile, under \textit{Asym.} and \textit{Inst.} noise, their post-correction Error Rates increase substantially, ranging from 23\% to 28\%. These results indicate their limited robustness to structured and instance-dependent label corruption.
By comparison, \tool maintains consistently strong performance across all noise types, with post-correction Error Rates remaining around 8.5\%. These results highlight the advantage of \tool over the baseline methods in handling realistic and complex noise patterns.

\subsubsection{RQ3: Component Analysis}
\label{sec:ablation_study}

% \begin{table*}\centering
% \caption{\centering RQ3: Impact of \tool's component on the overall performance}
% \label{tab:rq3}
% \begin{tabular}{l|l|rr|rr|rr|rrr}\toprule
% \multirow{2}{*}{} & &\multicolumn{2}{|c|}{\textbf{Agnews}} &\multicolumn{2}{|c|}{\textbf{CIFAR-10}} &\multicolumn{2}{|c}{\textbf{CIFAR-100}} & \multicolumn{2}{|c}{\textbf{Clickbait}}\\\midrule
% &Noise rate (\%) &\textbf{20.00} &\textbf{40.00} &\textbf{20.00} &\textbf{40.00} &\textbf{20.00} &\textbf{40.00} &\textbf{20.00} &\textbf{40.00} \\ \midrule

% \multirow{2}{*}{\textbf{$M_{clean}$}} 
% &Precision (↑) &71.60 &82.28 &84.21 &90.35 &25.27 &39.20 &XXX & XXX\\
% &Error rate (↓) &9.90 &18.90 &6.31 &9.64 &25.10 &39.80 & XXX & XXX\\\midrule
% \multirow{2}{*}{\textbf{$M_{clean}$ + $M_{noisy}$ }} &Precision (↑) &77.41 &89.47 &84.64 &91.28 &37.88 &45.78 & 96.12 & 97.76\\
% &Error rate (↓) &6.78 &14.00 &4.01&7.48 &20.26 &34.46 & 1.68 & 5.87\\
% \bottomrule
% \end{tabular}
% \end{table*}

%Please add the following packages if necessary:
%\usepackage{booktabs, multirow} % for borders and merged ranges
%\usepackage{soul}% for underlines
%\usepackage{xcolor,colortbl} % for cell colors
%\usepackage{changepage,threeparttable} % for wide tables
\begin{table}[!htp]\centering\small
\caption{\centering RQ3: Impact of \tool's component on the overall performance}
\label{tab:rq3}
\resizebox{\textwidth}{!}{ 
\begin{tabular}{l|r|r|r|r|rr}\toprule
& &\multicolumn{2}{c|}{$\mathcal{M}_{clean}$} &\multicolumn{2}{c}{$\mathcal{M}_{clean}$ + $\mathcal{M}_{noisy}$} \\\cmidrule{3-6}
&Noise rate (\%) &\textbf{Precision (↑)} &\textbf{Error Rate (↓)} &\textbf{Precision (↑)} &\textbf{Error Rate (↓)}  \\\midrule
\multirow{2}{*}{Agnews} &20 &71.60 &9.90 &77.41 &6.78 \\
&40 &82.28 &18.90 &89.47 &14.00 \\\midrule
\multirow{2}{*}{Clickbait} &20 &93.58 &6.40 &96.12 &1.68 \\
&40 &96.38 &6.10 &97.76 &5.87 \\\midrule
\multirow{2}{*}{CIFAR-10} &20 &84.21 &6.31 &84.64 &4.01 \\
&40 &90.35 &9.64 &91.28 &7.48 \\\midrule
\multirow{2}{*}{CIFAR-100} &20 &25.27 &25.10 &37.88 &20.26 \\
&40 &39.2 &39.80 &45.78 &34.46 \\\midrule
\multirow{2}{*}{CodeXGLUE} &20 &25.41 &30.80 &33.14 &19.85 \\
&40 &43.00 &40.40 &52.87 &36.44 \\
\bottomrule
\end{tabular}
}
\end{table}

\tool performs noisy label repair through Bayesian inference that integrates the predictions of two core components: the \textit{Clean Label Predictor} $M_{clean}$ and the \textit{Noisy Label Estimator} $M_{noise}$ (Fig.~\ref{fig:pipeline}). 
To examine the contribution of each component, we conduct an ablation study comparing the performance of \tool when using $\mathcal{M}_{clean}$ alone and when combining $\mathcal{M}_{clean}$ and $\mathcal{M}_{noise}$ within the Bayesian framework.
Table~\ref{tab:rq3} reports the experimental results across datasets under different noise levels.

Across all datasets and noise rates, \textit{incorporating both $\mathcal{M}_{clean}$ and $\mathcal{M}_{noise}$ consistently improves correction Precision and reduces Error Rate compared to relying solely on $\mathcal{M}_{clean}$.}
In particular, the integration of  $\mathcal{M}_{noise}$ yields relative improvements of up to 50\% in Precision and 19\% reduction in Error rate.
These results demonstrate that relying solely on the probabilities estimated by $\mathcal{M}_{clean}$ is insufficient for robust label repair, as it does not account for the underlying noise distribution.
By contrast, integrating the complementary signals from both $\mathcal{M}_{clean}$ and $\mathcal{M}_{noise}$ within a Bayesian framework enables more calibrated and reliable corrections, leading to substantially improved performance.

\subsubsection{RQ4: Downstream Utility}
%Please add the following packages if necessary:
%\usepackage{booktabs, multirow} % for borders and merged ranges
%\usepackage{soul}% for underlines
%\usepackage{xcolor,colortbl} % for cell colors
%\usepackage{changepage,threeparttable} % for wide tables
%If the table is too wide, replace \begin{table}[!htp]...\end{table} with
%\begin{adjustwidth}{-2.5 cm}{-2.5 cm}\centering\begin{threeparttable}[!htb]...\end{threeparttable}\end{adjustwidth}
\begin{table}[!htp]\centering
\caption{RQ4. Downstream model performance when training with dataset before (Original) and after label correction
}\label{tab:rq4}
%\resizebox{\textwidth}{!}{ % use this if the table is too large
\begin{tabular}{l|r|r|r|rr}\toprule
&\textbf{Original} &\textbf{\semiv} &\textbf{\semiv} &\textbf{\tool} \\\midrule
Accuracy &71.17 &73.07 &73.40 &75.50 \\
Weighted F1 &70.37 &71.47 &72.21 &74.43 \\
\bottomrule
\end{tabular}
\end{table}

Table~\ref{tab:rq4} shows the downstream classification performance of a Multilayer Perceptron (MLP) classifier trained on the original noisy Agnews dataset and on datasets repaired by different label correction methods.
As expected, \textit{improving label quality leads to better training signals and enhanced downstream performance.}
When trained on the original dataset, the classifier achieves an accuracy of 71.17\% and a weighted F1 score of 70.37\%. 
Applying \semiv and \semir for label correction yields improvements of approximately 3\% relative gain in accuracy. \tool achieves a larger improvement. By more precisely repairing corrupted labels, \tool increases the downstream accuracy to 75.50\%, corresponding to a relative gain of about 6\% over the original dataset.
These results underscore the importance of clean supervision for training reliable classification models. Compared to existing baselines, \tool more effectively recovers accurate labels, resulting in higher-quality training data and significantly improved generalization performance of downstream models.

\section{Related Work}

\textbf{Automated data annotation:}
Data annotation is essential yet costly for training high-quality ML/DL models~\cite{schmarje2022one}, which has motivated extensive research on automated and semi-automated labeling techniques~\cite{hou2024kgred,song2024optimal,zhu2024apt,galhotra2021adaptive}. While these approaches can substantially reduce annotation effort, prior studies have reported limitations in the accuracy and consistency of generated labels~\cite{llm-label-quality,chatgpt-label}.

\textbf{Corrupted label detection:}
Numerous approaches~\cite{simifeat,retrieval-based,yu2023delving,cola,sent,noiserank} have been proposed to identify corrupted labels.
For example, SENT~\cite{sent} leverages a small trusted subset to model noise distribution and uses it to distinguish corrupted labels. Retrieval-based approaches~\cite{retrieval-based} identify noisy labels by comparing an instance with its nearest neighbors. \cola~\cite{cola} further improves detection by modeling both local and global instance relationships. While effective at identifying suspicious labels, these methods focus primarily on detection and typically rely on discarding or reweighting noisy instances. 
\tool is complementary to these approaches, as it operates on detected corrupted labels and explicitly repairs them rather than removing them from the dataset.

\textbf{Corrupted label correction:}
Only a limited number of studies~\cite{simifeat,docta} have explicitly investigated corrupted label correction from a \textit{data-centric} perspective. \semi~\cite{simifeat,docta} detects unreliable annotations based on neighborhood disagreement and subsequently corrects them using heuristic strategies such as majority voting or ranking-based label replacement.
In contrast, the majority of existing work adopts a \textit{model-centric} perspective, focusing on designing learning algorithms that are inherently robust to noisy labels rather than explicitly repairing them~\cite{chenglearning,li2019learning,self-learning-with-noise,lienen2024mitigating}.
Different from these approaches, \tool directly targets the \textit{label repair} problem by explicitly modeling the relationship between clean labels, observed noisy labels, and input features. By formulating label correction as a Bayesian inference problem, \tool infers the most probable clean label for each instance, enabling systematic and principled correction of corrupted annotations. As a result, \tool complements model-centric noise-robust learning methods and provides a reusable, high-quality dataset that benefits subsequent learning tasks.
\section{Conclusion}

Label noise remains a critical challenge in building reliable ML/DL systems, as corrupted annotations can substantially degrade both model performance and robustness.  
Although recent data-centric methods have improved the detection of noisy labels, effective and reliable label correction methods remain relatively underexplored.
To address this gap, we introduce \tool, an end-to-end data-centric framework for corrupted label detection and correction. \tool jointly leverages local and global relationships among data instances to identify suspicious samples, and further repairs corrupted labels by estimating the most probable clean label based on both input features and observed noisy labels. Extensive experiments across diverse datasets, noise types, and noise rates demonstrate that \tool consistently improves label correction quality and enhances downstream model performance, while also complementing existing noise detection approaches.

\bibliographystyle{splncs04}
\bibliography{ref}

@article{bert,
  title={Bert: Pre-training of deep bidirectional transformers for language understanding},
  author={Devlin, Jacob},
  journal={arXiv preprint arXiv:1810.04805},
  year={2018}
}

@inproceedings{lienen2024mitigating,
  title={Mitigating label noise through data ambiguation},
  author={Lienen, Julian and H{\"u}llermeier, Eyke},
  booktitle={Proceedings of the AAAI Conference on Artificial Intelligence},
  volume={38},
  number={12},
  pages={13799--13807},
  year={2024}
}

@inproceedings{clip,
  title={Learning transferable visual models from natural language supervision},
  author={Radford, Alec and Kim, Jong Wook and Hallacy, Chris and Ramesh, Aditya and Goh, Gabriel and Agarwal, Sandhini and Sastry, Girish and Askell, Amanda and Mishkin, Pamela and Clark, Jack and others},
  booktitle={International conference on machine learning},
  pages={8748--8763},
  year={2021},
  organization={PMLR}
}

@inproceedings{retrieval-based,
  title={Retrieval-Based Unsupervised Noisy Label Detection on Text Data},
  author={Liu, Peiyang and Yang, Jinyu and Wang, Lin and Wang, Sen and Hao, Yunlai and Bai, Huihui},
  booktitle={Proceedings of the 32nd ACM International Conference on Information and Knowledge Management},
  pages={4099--4104},
  year={2023}
}

@inproceedings{noiserank,
  title={Noiserank: Unsupervised label noise reduction with dependence models},
  author={Sharma, Karishma and Donmez, Pinar and Luo, Enming and Liu, Yan and Yalniz, I Zeki},
  booktitle={European conference on computer vision},
  pages={737--753},
  year={2020},
  organization={Springer}
}

@inproceedings{sent,
  title={Learning to Detect Noisy Labels Using Model-Based Features},
  author={Wang, Zhihao and Lin, Zongyu and Wen, Junjie and Chen, Xianxin and Liu, Peiqi and Zheng, Guidong and Chen, Yujun and Yang, Zhilin},
  booktitle={Findings of the Association for Computational Linguistics: EMNLP 2022},
  pages={5796--5808},
  year={2022}
}

@article{survey,
  title={Learning from noisy labels with deep neural networks: A survey},
  author={Song, Hwanjun and Kim, Minseok and Park, Dongmin and Shin, Yooju and Lee, Jae-Gil},
  journal={IEEE transactions on neural networks and learning systems},
  volume={34},
  number={11},
  pages={8135--8153},
  year={2022},
  publisher={IEEE}
}

@inproceedings{self-learning-with-noise,
  title={Deep self-learning from noisy labels},
  author={Han, Jiangfan and Luo, Ping and Wang, Xiaogang},
  booktitle={Proceedings of the IEEE/CVF international conference on computer vision},
  pages={5138--5147},
  year={2019}
}

@inproceedings{chenglearning,
  title={Learning with Instance-Dependent Label Noise: A Sample Sieve Approach},
  author={Cheng, Hao and Zhu, Zhaowei and Li, Xingyu and Gong, Yifei and Sun, Xing and Liu, Yang},
  booktitle={International Conference on Learning Representations},
    year= {2021}
}

@inproceedings{xiao2015learning,
  title={Learning from massive noisy labeled data for image classification},
  author={Xiao, Tong and Xia, Tian and Yang, Yi and Huang, Chang and Wang, Xiaogang},
  booktitle={Proceedings of the IEEE conference on computer vision and pattern recognition},
  pages={2691--2699},
  year={2015}
}

@inproceedings{simifeat,
  title={Detecting corrupted labels without training a model to predict},
  author={Zhu, Zhaowei and Dong, Zihao and Liu, Yang},
  booktitle={International conference on machine learning},
  pages={27412--27427},
  year={2022},
  organization={PMLR}
}

@inproceedings{li2019learning,
  title={Learning to learn from noisy labeled data},
  author={Li, Junnan and Wong, Yongkang and Zhao, Qi and Kankanhalli, Mohan S},
  booktitle={Proceedings of the IEEE/CVF conference on computer vision and pattern recognition},
  pages={5051--5059},
  year={2019}
}

@inproceedings{yu2023delving,
  title={Delving into noisy label detection with clean data},
  author={Yu, Chenglin and Ma, Xinsong and Liu, Weiwei},
  booktitle={International Conference on Machine Learning},
  pages={40290--40305},
  year={2023},
  organization={PMLR}
}

@inproceedings{kim2024learning,
  title={Learning Discriminative Dynamics with Label Corruption for Noisy Label Detection},
  author={Kim, Suyeon and Lee, Dongha and Kang, SeongKu and Chae, Sukang and Jang, Sanghwan and Yu, Hwanjo},
  booktitle={Proceedings of the IEEE/CVF Conference on Computer Vision and Pattern Recognition},
  pages={22477--22487},
  year={2024}
}

@article{gao2016resistance,
  title={On the resistance of nearest neighbor to random noisy labels},
  author={Gao, Wei and Yang, Bin-Bin and Zhou, Zhi-Hua},
  journal={arXiv preprint arXiv:1607.07526},
  year={2016}
}

@inproceedings{wang2024human,
  title={Human-LLM collaborative annotation through effective verification of LLM labels},
  author={Wang, Xinru and Kim, Hannah and Rahman, Sajjadur and Mitra, Kushan and Miao, Zhengjie},
  booktitle={Proceedings of the CHI Conference on Human Factors in Computing Systems},
  pages={1--21},
  year={2024}
}

@article{cola,
  title={Leveraging local and global relationships for corrupted label detection},
  author={Lam, Phong and Nguyen, Ha-Linh and Dang, Xuan-Truc Dao and Tran, Van-Son and Le, Minh-Duc and Nguyen, Thu-Trang and Nguyen, Son and Vo, Hieu Dinh},
  journal={Future Generation Computer Systems},
  pages={107729},
  year={2025},
  publisher={Elsevier}
}

@inproceedings{zhu2024apt,
  title={APT-Pipe: A Prompt-Tuning Tool for Social Data Annotation using ChatGPT},
  author={Zhu, Yiming and Yin, Zhizhuo and Tyson, Gareth and Haq, Ehsan-Ul and Lee, Lik-Hang and Hui, Pan},
  booktitle={Proceedings of the ACM on Web Conference 2024},
  pages={245--255},
  year={2024}
}

@article{sharifani2023machine,
  title={Machine learning and deep learning: A review of methods and applications},
  author={Sharifani, Koosha and Amini, Mahyar},
  journal={World Information Technology and Engineering Journal},
  volume={10},
  number={07},
  pages={3897--3904},
  year={2023}
}

@article{docta,
  title={Unmasking and improving data credibility: A study with datasets for training harmless language models},
  author={Zhu, Zhaowei and Wang, Jialu and Cheng, Hao and Liu, Yang},
  journal={arXiv preprint arXiv:2311.11202},
  year={2023}
}

@article{schmarje2022one,
  title={Is one annotation enough?-a data-centric image classification benchmark for noisy and ambiguous label estimation},
  author={Schmarje, Lars and Grossmann, Vasco and Zelenka, Claudius and Dippel, Sabine and Kiko, Rainer and Oszust, Mariusz and Pastell, Matti and Stracke, Jenny and Valros, Anna and Volkmann, Nina and others},
  journal={Advances in Neural Information Processing Systems},
  volume={35},
  pages={33215--33232},
  year={2022}
}

@article{hou2024kgred,
  title={KGRED: Knowledge-graph-based rule discovery for weakly supervised data labeling},
  author={Hou, Wenjun and Hong, Liang and Zhu, Ziyi},
  journal={Information Processing \& Management},
  volume={61},
  number={5},
  pages={103816},
  year={2024},
  publisher={Elsevier}
}

@inproceedings{galhotra2021adaptive,
  title={Adaptive rule discovery for labeling text data},
  author={Galhotra, Sainyam and Golshan, Behzad and Tan, Wang-Chiew},
  booktitle={Proceedings of the 2021 International conference on management of data},
  pages={2217--2225},
  year={2021}
}

@article{song2024optimal,
  title={Optimal block-wise asymmetric graph construction for graph-based semi-supervised learning},
  author={Song, Zixing and Zhang, Yifei and King, Irwin},
  journal={Advances in Neural Information Processing Systems},
  volume={36},
  year={2024}
}

@article{llm-label-quality,
  title={Testing the reliability of chatgpt for text annotation and classification: A cautionary remark},
  author={Reiss, Michael V},
  journal={arXiv preprint arXiv:2304.11085},
  year={2023}
}

@inproceedings{chatgpt-label,
  title={Is chatgpt better than human annotators? potential and limitations of chatgpt in explaining implicit hate speech},
  author={Huang, Fan and Kwak, Haewoon and An, Jisun},
  booktitle={Companion proceedings of the ACM web conference 2023},
  pages={294--297},
  year={2023}
}

@misc{website, 
title = {A data-centric framework for detecting and correcting corrupted labels},
author={ Nguyen, Ha-Linh and Nguyen, Hong-Anh and La, Minh-Duc and Nguyen, Thu-Trang and Nguyen, Son and Vo, Dinh Hieu},
url={https://github.com/iSE-UET-VNU/RELABELER}
}

\end{document}